\documentclass[10pt,twocolumn,letterpaper]{article}

\usepackage[pagenumbers]{cvpr} 

\definecolor{cvprblue}{rgb}{0.21,0.49,0.74}
\usepackage[pagebackref,breaklinks,colorlinks,allcolors=cvprblue]{hyperref}
\usepackage{float}
\usepackage{hyperref}
\usepackage{url}
\usepackage{algorithm} 
\usepackage{algorithmic}
\usepackage{graphicx}
\usepackage{booktabs}
\usepackage{multirow} 
\usepackage{makecell}

\def\paperID{12771} 
\def\confName{CVPR}
\def\confYear{2026}

\title{Fine-Grained GRPO for Precise Preference Alignment in Flow Models}

\author{\textbf{Yujie Zhou}$^{1,4*}$\quad
\textbf{Pengyang Ling}$^{2,4*}$\quad
\textbf{Jiazi Bu}$^{1,4*}$\quad
\textbf{Yibin Wang}$^{3,5}$\\
\textbf{Yuhang Zang}$^{4}$\quad 
\textbf{Jiaqi Wang}$^{4,5\dag}$\quad
\textbf{Li Niu}$^{1\dag}$\quad
\textbf{Guangtao Zhai}$^{1}$\\
\textsuperscript{\rm 1}Shanghai Jiao Tong University \
\textsuperscript{\rm 2}University of Science and Technology of China\\
\textsuperscript{\rm 3}Fudan University \quad
\textsuperscript{\rm 4}Shanghai AI Laboratory \quad
\textsuperscript{\rm 5}Shanghai Innovation Institute \\
\url{https://bujiazi.github.io/g2rpo.github.io/} 
} 

\begin{document}


\twocolumn[{
\maketitle
\begin{center}
    \centering
    \captionsetup{type=figure}
    \includegraphics[width=\textwidth]{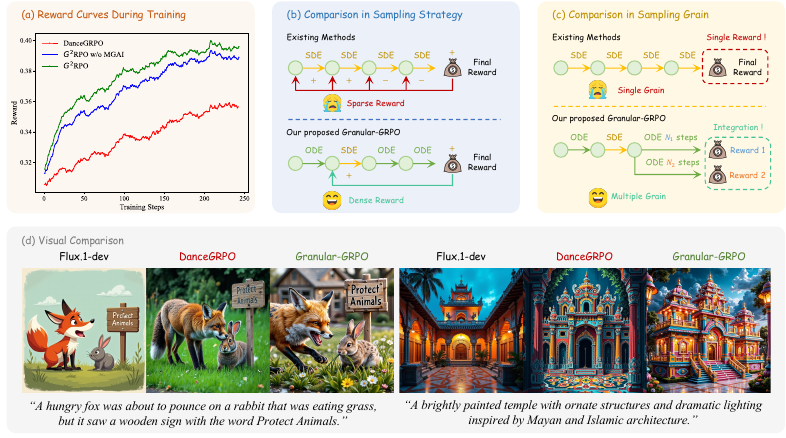}
    \caption{ \textbf{Comparison between our G$^2$RPO and existing studies}. 
    (a) G$^2$RPO outperforms DanceGRPO in reward scores (HPS-v2.1 in this figure).
    (b) Sampling strategy comparison. G$^2$RPO acquire a dense reward by confining stochasticity to individual sampling steps. 
    (c) Sampling grain comparison. G$^2$RPO achieves a comprehensive evaluation of each sampling direction by integrating advantages from multi-granularity ODE denoising. 
    (d) Visual Comparison. Compared to the baseline method, the images generated by G$^2$RPO are more aligned with human preferences.
        }
    \label{fig:teaser}
\end{center}
}]

{
  \renewcommand{\thefootnote}{\fnsymbol{footnote}}
  \footnotetext[1]{Equal contribution. \textsuperscript{\dag} Corresponding authors.}
}

\begin{abstract}
The incorporation of online reinforcement learning (RL) into diffusion and flow-based generative models has recently gained attention as a powerful paradigm for aligning model behavior with human preferences. By leveraging stochastic sampling via Stochastic Differential Equations (SDEs) during the denoising phase, these models can explore a variety of denoising trajectories, enhancing the exploratory capacity of RL. However, despite their ability to discover potentially high-reward samples, current approaches often struggle to effectively align with preferences due to the sparsity and narrowness of reward feedback. To overcome this limitation, we introduce a novel framework called Granular-GRPO \textbf{(G$^2$RPO)}, which enables fine-grained and comprehensive evaluation of sampling directions in the RL training of flow models. Specifically, we propose a \textbf{Singular Stochastic Sampling} mechanism that supports step-wise stochastic exploration while ensuring strong correlation between injected noise and reward signals, enabling more accurate credit assignment to each SDE perturbation. Additionally, to mitigate the bias introduced by fixed-granularity denoising, we design a \textbf{Multi-Granularity Advantage Integration} module that aggregates advantages computed across multiple diffusion scales, resulting in a more robust and holistic assessment of sampling trajectories. Extensive experiments on various reward models, including both in-domain and out-of-domain settings, demonstrate that our G$^2$RPO outperforms existing flow-based GRPO baselines, highlighting its effectiveness and generalization capability.
\end{abstract}    
\vspace{-1em}
\section{Introduction}
\label{sec:intro}
Recent advances in generative models, particularly diffusion models~\citep{ho2020denoising,song2020denoising,song2020score} and flow models~\citep{lipman2022flow,liu2022flow,peebles2023scalable},
have revolutionized visual content creation,
offering unprecedented capabilities in generating high-quality images~\citep{rombach2022high,podell2023sdxl,esser2024scaling,flux2024} and videos~\citep{blattmann2023stable,chen2024videocrafter2,guo2023animatediff,kong2024hunyuanvideo,wan2025wan}. 
However, a key challenge remains in
aligning model outputs with the diverse and complex human preferences.
To tackle this challenge, reinforcement learning from human feedback (RLHF)~\citep{fan2023dpok, black2023training}
has emerged as a promising solution,
characterized by its adaptability and cost-effectiveness.
Paradigms such as Proximal Policy Optimization (PPO)~\citep{schulman2017proximal}, 
Direct Policy Optimization (DPO)~\citep{rafailov2023direct}, and Group Relative Policy Optimization (GRPO)~\citep{shao2024deepseekmath} 
have been introduced. 
Among these, GRPO stands out as an innovative online reinforcement learning approach.
Leveraging group comparisons to optimize policies, 
GRPO eliminates the need for a separate value model, 
achieving greater flexibility and scalability.


To integrate GRPO into flow-based generative models, 
Flow-GRPO~\citep{liu2025flow} and DanceGRPO~\citep{xue2025dancegrpo}
substitute the deterministic ODE sampler with an SDE formulation,
wherein the injected stochasticity deliberately perturbs the denoising direction at each step.
Although the resulting samples enable exhaustive per-step exploration for reinforcement learning,  
they simultaneously underscore the difficulty of attributing the final reward to any specific random perturbation, 
thereby constraining model trainability.
Specifically, most existing flow-based GRPO methods 
encounter two core issues in evaluating group denoising directions:
1) \textbf{Sparse reward}: 
As shown in Fig.~\ref{fig:teaser} (b), the final reward signal is uniformly assigned to each SDE sampling step, 
which cannot be precisely aligned with the sampling direction at each step, 
leading to inaccuracies for the optimization at individual steps.
2) \textbf{Incomplete evaluation}: 
While the same single-step SDE sampling direction can
yield different rewards under varying denoising intervals, 
an ideal high-quality direction (Sample 2 and Sample 3 shown in Fig.~\ref{fig:obs}) intuitively is robust to the granularity of the remaining sampling trajectory.
This means that although the absolute quality of the final results may vary with the number of steps, 
an excellent denoising direction still maintains a superior relative position within the group.
Nevertheless, as shown in Fig.~\ref{fig:teaser}(c), 
existing methods suffer from the drawback that each denoising direction is 
rigidly associated with a fixed number of denoising steps.
Consequently, it leads to a single granularity of the denoised images, 
making it impossible to conduct a thorough comparison across the group.


To address these limitations, 
we propose \textbf{G}ranular-\textbf{GRPO} (G$^2$RPO), 
a novel online reinforcement learning framework specifically
designed for precise and comprehensive reward signals.
First, mirroring the sparse-reward problem~\citep{hare2019dealing, liang2024step} that plagues RLHF,
the reward signal in the SDE sampling process is delivered only after an entire sequence of decisions.
This long delay undermines the credit-assignment chain,
preventing the linking of the terminal reward with any specific earlier 
action and thereby inducing sluggish, unstable learning. 
Therefore, we propose a simple yet effective sampling strategy,
termed \textbf{Singular Stochastic Sampling}.
As illustrated in Fig~\ref{fig:teaser} (b),
this strategy applies the SDE formulation at a single time step to generate a group of denoising directions, 
while employing deterministic ODE sampling for all other steps.
By concentrating stochasticity at one specific step, the proposed method establishes a strong correlation between 
the reward signal and the injected noise, enabling stable model optimization. 
Secondly, we propose a \textbf{Multi-Granularity Advantage Integration} (MGAI) module.
As depicted in Fig.~\ref{fig:teaser}(c),
instead of binding each denoising direction to a fixed subsequent denoising granularity, 
the denoising directions in the same group are assigned to a spectrum of denoising steps, producing images with different granularities.
The corresponding reward signals of these images are then fused into a unified advantage estimate, yielding a comprehensive evaluation of the current state’s value.
With the support of the Singular Stochastic Sampling strategy and the Multi-Granularity Advantage Integration module,
G$^2$RPO can provide a more precise and comprehensive reward signal, 
thereby enhancing the upper limit of the GRPO model training. 
As shown in Fig~\ref{fig:teaser} (a), our reward curves exhibit stable 
improvements over the baseline during training. Additionally, 
Fig~\ref{fig:teaser} (d) illustrates the images generated by G$^2$RPO, 
highlighting its advantages in text prompt adherence and detail fidelity.

Our contributions can be summarized as follows: 
\begin{itemize}
\item \textbf{Granular-GRPO}: A novel flow-based GRPO framework designed to provide a precise and comprehensive 
evaluation of the denoising directions sampled by the SDE, thereby improving the precision of model optimization. 
\item \textbf{Singular Stochastic Sampling}: A sampling strategy confines stochasticity to individual sampling steps, 
addressing the sparse reward issue associated with long-range stochasticity injection.
\item \textbf{Multi-Granularity Advantage Integration}: 
A module integrates the advantages of multi-granularity denoised images and 
enables a comprehensive evaluation of each sampling direction.
\end{itemize}
\section{Related Work}
\subsection{Alignment for Large Language Models}
Recent years have witnessed a paradigm shift from supervised fine-tuning~\citep{dong2023abilities, sun2024supervised} 
to online reinforcement learning~\citep{shani2024multi, abdulhai2023lmrl, jaech2024openai} when aligning Large Language Models (LLMs) with human intent,
which is known as Reinforcement Learning from Human Feedback (RLHF)~\citep{bai2025qwen2, ouyang2022training}. 
Early RLHF pipelines typically involve training a reward model from pairwise comparisons to
predict human preferences and guide a policy model through reinforcement learning algorithms 
like Proximal Policy Optimization (PPO)~\citep{schulman2017proximal}.
Despite their effectiveness, PPO introduces intensive computational overhead due to its 
iterative optimization process and the need for frequent interactions with the environment.
To address these computational challenges, value-free alternatives have been proposed,
such as Group Relative Policy Optimization (GRPO)~\citep{shao2024deepseekmath}. 
Adopted by leading LLM DeepSeek-R1~\citep{guo2025deepseek},  
GRPO aims to optimize policies based on relative preferences within a group of samples, 
providing a robust signal for policy improvement, particularly when absolute rewards are difficult to define or noisy. 
These advancements in LLM alignment provide a strong foundation for exploring similar human-centric optimization strategies in visual generation domains.

\subsection{Alignment for Flow Models}
Diffusion and Flow models~\citep{ho2020denoising,song2020denoising,song2020score,peebles2023scalable,rombach2022high}, which offer flexible visual creation through an iterative denoising process, have revolutionized the field of visual synthesis and become a pivotal part of generative models.
Building on the success of aligning LLMs with human preferences, similar techniques have recently been transplanted to diffusion and flow models~\citep{podell2023sdxl,esser2024scaling,flux2024}. 
Pioneer works like DDPO~\citep{black2023training} and ReFL~\citep{xu2023imagereward} apply PPO to finetune diffusion models for improved aesthetic performance and human feedback alignment. These methods face challenges inherent to RL, including high variance, low efficiency, and sparse reward. Diffusion-DPO~\citep{wallace2024diffusion} adapts the Direct Preference Optimization (DPO)~\citep{rafailov2023direct} framework to directly optimize diffusion models from paired preference data, bypassing the need for an explicit reward model but suffering from distribution shift since no new samples are collected during training. Recent efforts such as DanceGRPO~\citep{xue2025dancegrpo} and Flow-GRPO~\citep{liu2025flow} enable GRPO-style policy updates by converting the ODE sampling into an equivalent SDE to each timestep,
thereby acquiring a group of denoising directions for statistical sampling and RL exploration for flow models. More recently, MixGRPO~\citep{li2025mixgrpo} has improved training efficiency through a hybrid ODE-SDE sampling approach while maintaining comparable performance. However, these methods are generally constrained by sparse rewards due to long-range stochasticity injection and the binding of each sampling direction to a fixed denoising granularity. These paradigms restrict the ability to conduct a comprehensive evaluation of each sampling direction, limiting the optimization ceiling of GRPO training.
\section{Granular-GRPO}
\label{sec:method}
\begin{figure*}[t]
    \centering
    \includegraphics[width=\textwidth]{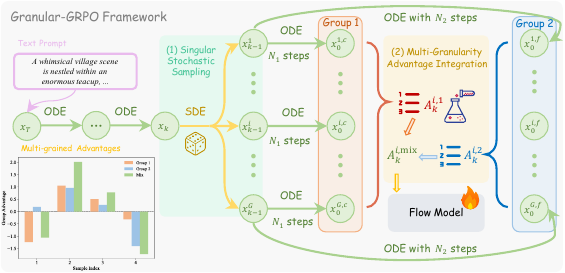}
    \caption{ \textbf{Overview of G$^2$RPO}. 
    Given a text prompt and an initial noise, our Singular Stochastic Sampling strategy
    employs SDE sampling solely at a single step and samples a group of distinct denoising directions.
    Then, the Multi-Granularity Advantage Integration module executes multi-granularity ODE denoising for each direction
    and integrates the advantages to produce a comprehensive evaluation for each sampling direction.
    For simplicity, the figure shows one coarse-grained path (denoted as $c$) and one fine-grained path (denoted as $f$).
    }
    \label{fig:pipeline}
\end{figure*}

As an online RL algorithm, 
flow-based GRPO methods utilize ODE-to-SDE conversion with the original model’s marginal distribution 
to sample a group of denoising directions for optimization.
A core issue underlying this paradigm is to obtain precise and comprehensive assessments of each sampling direction.
To this end, we introduce the Granular-GRPO framework to 
(i) confine stochasticity to individual steps (Singular Stochastic Sampling) for more precise reward signals, and 
(ii) integrate the advantages derived from multi-granularity denoising results 
(Multi-Granularity Advantage Integration) to acquire a more comprehensive evaluation, as shown in Fig.~\ref{fig:pipeline}.

\subsection{Preliminary}
For the flow-based GRPO methods~\citep{xue2025dancegrpo, liu2025flow, li2025mixgrpo}, the denoising process is first modeled as a multi-step Markov decision process (MDP).
Given a prompt $\boldsymbol{c}$, the agent with a flow model $p_\theta$ produce a reverse-time trajectories defined as 
$\Gamma=\left(\mathbf{s}_T, \mathbf{a}_T, \mathbf{s}_{T-1}, \mathbf{a}_{T-1}, \ldots, \mathbf{s}_0, \mathbf{a}_0\right)$,
where $\mathbf{s}_t = (\boldsymbol{c}, t , \boldsymbol{x}_t)$ is the state at timestep $t$ and $\boldsymbol{x}_t$ is the corresponding noisy sample.
Specifically, $\boldsymbol{x}_T \sim N(0, I)$ and $\boldsymbol{x}_0$ is the denoised image.
The action $\mathbf{a}_t$ represents the single step denoising process with the policy $\pi_{\theta}$,
indicating a sampling direction $\frac{d\boldsymbol{x}}{dt}$ from $\boldsymbol{x}_t$ to $\boldsymbol{x}_{t-1}$,
i.e. $\boldsymbol{x}_{t-1} \sim \pi_{\theta}(\boldsymbol{x}_{t-1}| \boldsymbol{x}_t, \boldsymbol{c})$.

\noindent \textbf{SDE Sampling.} As an online RL algorithm, GRPO needs grouped outputs and relative advantages to optimize policies.
However, the flow matching model utilizes a deterministic ODE to sample the denoising direction:
\begin{equation}
    \label{eq:ode}
    d \boldsymbol{x}_t=\boldsymbol{v}_\theta(\boldsymbol{x}_t,t) d t,
\end{equation}
where, $\boldsymbol{v}_\theta(\boldsymbol{x}_t,t)$ is the model output, given the noisy sample $\boldsymbol{x}_t$ and timestep $t$.

To match GRPO's stochastic sampling requirements, Flow-GRPO~\citep{liu2025flow} converts the ODE into an equivalent SDE sampling with 
the same marginal distribution:
\begin{equation}
d \boldsymbol{x}_t=(\boldsymbol{v}_\theta(\boldsymbol{x}_t, t)+
\frac{\sigma_t^2}{2 t}(\boldsymbol{x}_t+(1-t) \boldsymbol{v}_\theta(\boldsymbol{x}_t, t))) d t+\sigma_t d \boldsymbol{w}_t,
\end{equation}
where $d\boldsymbol{w}_t$ denotes Wiener process increments, and $\sigma_t$ controls the stochasticity injected into the sampling direction.
Furthermore, it can be discretized via the Euler–Maruyama scheme:
\begin{align}
\label{eq:euler}
\boldsymbol{x}_{t+\Delta t}
&= \boldsymbol{x}_t + \bigl(\boldsymbol{v}_\theta(\boldsymbol{x}_t, t)
    + \tfrac{\sigma_t^2}{2t}(\boldsymbol{x}_t + (1 - t)\boldsymbol{v}_\theta(\boldsymbol{x}_t, t))\bigr)
    \Delta t \nonumber \\
&\quad +\sigma_t\sqrt{\Delta t}\,\epsilon,
    \quad \epsilon \sim \mathcal{N}(0,I).
\end{align}
As defined in Flow-GRPO, $\sigma_t=\eta \sqrt{\frac{t}{1-t}}$, where the noise level is controlled by hyperparameter $\eta$.

\noindent \textbf{GRPO Training.} 
With SDE sampling, flow-based GRPO methods introduce stochasticity at each timestep to generate a group of $G$ images $\{\boldsymbol{x}_0^i\}^G_{i=1}$.
Then the reward model assigns a score $R(\boldsymbol{x}^i_0, \boldsymbol{c})$ to $\boldsymbol{x}_0^i$, and the advantage is computed as:
\begin{equation}
    \label{eq:adv}
    A_t^i=\frac{R(\boldsymbol{x}_0^i, \boldsymbol{c})-\operatorname{mean}(\{R(\boldsymbol{x}_0^j, \boldsymbol{c})\}_{j=1}^G)}{\operatorname{std}(\{R(\boldsymbol{x}_0^j, \boldsymbol{c})\}_{j=1}^G)} .
\end{equation}
Note that the advantages $A_0^i$ obtained from the final step image are uniformly broadcast to each step $A_t^i$ to evaluate the SDE sampling directions.
Finally, the policy model is optimized by maximizing the following objective:
\begin{equation}
    \label{eq:org_j}
    \mathcal{J}_{\text {Flow-GRPO }}(\theta)=\mathbb{E}_{\boldsymbol{c} \sim \mathcal{C},\left\{\boldsymbol{x}^i\right\}_{i=1}^G \sim \pi_{\theta_{\text {old }}}(\cdot \mid \boldsymbol{c})} f(r, A, \theta, \varepsilon, \beta),
\end{equation}
where
\begin{align}
\label{eq:f}
f(r, A, \theta, \varepsilon, \beta)=\frac{1}{G} \sum_{i=1}^G \frac{1}{T} \sum_{t=0}^{T-1}(\min (r_t^i(\theta) A_t^i, \nonumber\\
\operatorname{clip}(r_t^i(\theta), 1-\varepsilon, 1+\varepsilon) A_t^i)-\beta D_{\mathrm{KL}}(\pi_\theta \| \pi_{\mathrm{ref}}),
\end{align}
\begin{equation}
\label{eq:ratio}
r_t^i(\theta) =\frac{p_\theta\left(\boldsymbol{x}_{t-1}^i \mid \boldsymbol{x}_t^i, \boldsymbol{c}\right)}{p_{\theta_{\text {old }}}\left(\boldsymbol{x}_{t-1}^i \mid \boldsymbol{x}_t^i, \boldsymbol{c}\right)}.    
\end{equation}
Notably, $\beta$ is the hyperparameter that controls the proportion of KL loss.
Following the practices of DanceGRPO and MixGRPO, $\beta=0$ to achieve a stable training process.

\subsection{Singular Stochastic Sampling}
Traditional flow-based GRPO methods introduce ODE-to-SDE conversion sampling at every step
to inject stochasticity and uniformly assign the final image reward to each step's sampling direction. 
According to Eq.~\ref{eq:adv}, the advantage $A_t^i$ of each sampling direction at step $t$ is equally assigned with $A_0^{i}$.
Notably, the reward signal available only after multiple decision steps impedes the model's capability
to link the final reward to each decision, thereby resulting in imprecise and sparse rewards.

In accordance with the design of DanceGRPO, we define $M$ as the first half of the denoising timesteps,
i.e., \( M = \{T, T-1, \ldots, \lfloor T/2 \rfloor\} \).
This selection prioritizes the early denoising steps,
where the exploration space of the SDE is larger and significantly determines
the overall direction of the entire denoising chain.
In contrast, the improvements from GRPO in the later stages are relatively minor.
Consequently, training is conducted exclusively on these timesteps after each round of group image sampling, 
thereby optimizing computational efficiency.
Moreover, to acquire a dense reward for each SDE sampling direction,
a simple yet effective strategy is to confine the stochasticity to the single step selected for optimization. 
As shown in Fig.~\ref{fig:pipeline},
given a prompt $\boldsymbol{c}$ and initial noise $\boldsymbol{x}_T$, 
for each timestep $k \in M$,
a common starting point $\boldsymbol{x}_k$ for the group is acquired using ODE sampling from Eq.~\ref{eq:ode}.
Then, our Singular Stochastic Sampling strategy employs SDE sampling 
only at $\boldsymbol{x}_k$ and samples $G$ distinct denoising directions 
to get the next noisy state $ \{\boldsymbol{x}_{k-1}^i\}^G_{i=1}$. 
Each $\boldsymbol{x}_{k-1}^i$ undergoes $k-1$ steps of ODE sampling 
to generate a deterministic denoised image $\boldsymbol{x}_{0\gets k}^i$.
Based on our sampling strategy, 
the variance of the group reward $ \{R(\boldsymbol{x}_{0 \gets k}^i,\boldsymbol{c})\}^G_{i=1}$
is entirely determined by the distinct denoising directions
introduced by the SDE sampling at step $k$.
A step-aware, precise advantage can be acquired:
\begin{equation}
    A_k^i=\frac{R(\boldsymbol{x}_{0\gets k}^i, \boldsymbol{c})-\operatorname{mean}(\{R(\boldsymbol{x}_{0 \gets k}^i,
    \boldsymbol{c})\}_{i=1}^G)}{\operatorname{std}(\{R(\boldsymbol{x}_{0 \gets k}^i, \boldsymbol{c})\}_{i=1}^G)} .
\end{equation}
Consequently, the $f(r, A, \theta, \varepsilon, \beta)$ in Eq~\ref{eq:org_j} can be formulated as:
\begin{align}
\label{eq:sss}
f(r, A, \theta, \varepsilon, \beta) = \frac{1}{G} \sum_{i=1}^G \frac{1}{K}
\sum_{k \in M} \biggl(\min \Bigl( \nonumber r_k^i(\theta) A_k^i, \\
\operatorname{clip}\bigl( r_k^i(\theta), 1-\varepsilon, 1+\varepsilon \bigr) A_k^i \Bigr) \biggr).
\end{align}

\begin{figure*}[ht]
    \centering
    \includegraphics[width=\textwidth]{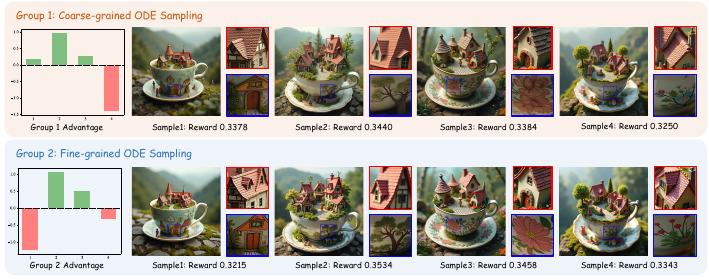}
    \caption{ \textbf{Visual Comparison of Images Denoised at Different Granularities.}
    Images denoised at different granularities exhibit variations in fine details and textures, 
    leading to inconsistent scoring by the Reward Model (HPS-v2.1).
    This observation reveals the insufficiency of a single-granularity evaluation of group advantage.
            }
    \label{fig:obs}
\end{figure*}

After the sampling phase is completed, the training of GRPO requires the computation of
$p_\theta\left(\boldsymbol{x}_{k-1}^i \mid \boldsymbol{x}_k^i, \boldsymbol{c}\right)$ to obtain $r_k^i(\theta)$ refer to eq~\ref{eq:ratio}.
In practice, each distinct sampling starting point $\boldsymbol{x}_k^i$ needs to be fed into the flow model 
to compute the corresponding ODE denoising direction $\boldsymbol{v}_k^i$.
However, our sampling strategy shares a common starting point $\boldsymbol{x}_k$,
allowing a group of $G$ samples to reuse the same $\boldsymbol{v}_k$, which in turn improves training efficiency. 

\subsection{Multi-Granularity Advantage Integration}

Singular Stochastic Sampling accurately constrains stochasticity into the single SDE step, 
ensuring a strong correlation between the reward and the injected noise. 
Then, how to acquire a comprehensive reward signal for the denoising direction of the current step
still requires further investigation.


As shown in Fig.~\ref{fig:obs}, we observe that under identical $\boldsymbol{x}_k$ and prompt $\boldsymbol{c}$ conditions,
the denoising trajectory generated by singular stochastic sampling is not robust when assessing the 
corresponding SDE denoising direction.
Specifically, while the overall content of the images generated from different denoising trajectories remains similar,
the details vary significantly due to the differing denoising intervals. 
This variation in denoising granularity leads to discrepancies in the scores assigned by the reward model,
which in turn influences the numerical values of the advantage within the group and even the optimization direction.

To conduct a comprehensive assessment of the single-step SDE sampling direction,
we propose a Multi-Granularity Advantage Integration module, which performs
multi-granularity denoising on the sampled denoising directions within a group. 
The advantages of the images denoised at different granularities are then integrated to form the final evaluation.
Specifically, as shown in Fig.~\ref{fig:pipeline}, step $k$ is the SDE sampling step, and 
$G$ distinct denoising directions are sampled to acquire next noisy state  $ \{\boldsymbol{x}_{k-1}^i\}^G_{i=1} $.
Under the conventional granularity condition, each $\boldsymbol{x}_{k-1}^i$ undergoes $k-1$ steps to obtain the final
denoised image.
The sequence of denoising timesteps can be represented as: $\mathcal{S}=\{1,2, \ldots, k-1\}$.
For our Multi-Granularity denoising module,
a set of integer scaling factors $\Lambda = \{\lambda_1, \lambda_2, ..., \lambda_j\}, |\Lambda|=J$ is defined to
represent different denoising granularities.
Each $\lambda_j$ implements interval sampling for different denoising granularity, 
that means sample every $\lambda_j$-th step from the total $k-1$ steps.
The denoising timestep sequence $\mathcal{S}_j$ can be formally represented as:
\begin{equation}
    \mathcal{S}_j=\{1,1+\lambda_j, 1+2 \lambda_j, \ldots,\left\lceil\frac{k-1}{\lambda_j}\right\rceil \lambda_j\},
\end{equation}
Our interval sampling approach ensures that the denoising process is performed at regular intervals defined by $\lambda_j$.
As $\lambda_j$ increases, the granularity becomes coarser,
allowing for a more flexible and adaptive denoising process.
For ease of illustration, $J=2$ in the Fig.~\ref{fig:pipeline}.

After $N_j$ subsequent steps denoising for $\{\boldsymbol{x}_{k-1}^i\}^G_{i=1}$, 
a group of noise-free images $\{x_0^{i,j}\}^G_{i=1}$ are generated.
Subsequently, different groups images gets the reward $\{R(\boldsymbol{x}_{0 \gets k}^{i,j},\boldsymbol{c})\}^G_{i=1}$ from a reward model 
and then compute the intra-group advantages $\{A_t^{i,j}\}^G_{i=1}$ with Eq.~\ref{eq:adv}.
Similar to the joint training with multiple reward models (e.g., HPS-v2.1 and CLIP Score) in DanceGRPO, 
where the advantages from different reward models are directly summed to provide 
a multi-dimensional comprehensive evaluation of a single denoising direction,
we combines the advantages from different granularities to get $\{A_{t}^{i,\text{mix}}\}^G_{i=1}$:
\begin{equation}
    A_{t}^{i,\text{mix}} = \sum_j^J A_t^{i,j}
\end{equation}
Accordingly, $f(r, A, \theta, \varepsilon, \beta)$ in Eq~\ref{eq:sss} can be updated to the following formula:
\begin{align}
f(r, A, \theta, \varepsilon, \beta)=\frac{1}{G} \sum_{i=1}^G \frac{1}{K} \sum_{k \in M}(\min (r_k^i(\theta) A_k^{i,\text{mix}}, \nonumber\\
\operatorname{clip}\left(r_k^i(\theta), 1-\varepsilon, 1+\varepsilon\right) A_k^{i,\text{mix}})),
\end{align}

Finally, our G$^2$RPO uses Eq.~\ref{eq:org_j} to optimize the policy $\pi_\theta$ across all timesteps in the set $M$.
A detailed algorithm is illustrated in Algorithm~\ref {algo:grpo}.
\begin{figure}[H]
\centering
\resizebox{\linewidth}{!}{ 
\begin{minipage}{\linewidth} 
\begin{algorithm}[H]
\caption{Granular-GRPO Training Process}
\label{algo:grpo}

\newcommand{\funccommd}[1]{{\textcolor{blue}{#1}}}
\newcommand{\mycommfont}[1]{{\scriptsize\itshape\textcolor{red}{#1}}}

\begin{algorithmic}[1]
    \STATE {\bfseries Require:} Prompt dataset $\mathcal{C}$, policy model $\pi_\theta$, reward model $R$, total sampling steps $T$
    \STATE {\bfseries Require:} SDE sampling timestep set $M$, Denoising granularities set $\Lambda$ ($|\Lambda|=J$)
    \FOR{training iteration $e=1$ {\bfseries to} $E$}
        \STATE Update old policy model: $\pi_{\theta_{\text{old}}} \gets \pi_\theta$
        \STATE Sample batch prompts $\mathcal{C}_b \sim \mathcal{C}$
        \FOR{prompt $\mathbf{c} \in \mathcal{C}_b$}
            \STATE Init same noise $\boldsymbol{x}_T \sim \mathcal{N}(0,\mathbf{I})$
            \FOR{$k \in M$}
                \FOR{$t=T$ to $0$}
                    \IF{$t > k$}
                        \STATE ODE Sampling: $\boldsymbol{x}_{t-1}$
                    \ELSIF{$t == k$}
                        \STATE SDE Sampling a group samples: \(\boldsymbol{x}_{k-1}^i\)
                    \ELSIF{$t < k$}
                        \FOR{$\lambda_j \in \Lambda$}
                            \STATE ODE Sampling with granularity $\lambda_j$: $\boldsymbol{x}_{t-1}^{i,j}$
                        \ENDFOR
                    \ENDIF
                \ENDFOR
                \STATE Get a group of reward: $R(\boldsymbol{x}_{0 \gets k}^{i,j})$
                \STATE $A_k^{j} \gets \frac{R(\boldsymbol{x}_{0 \gets k}^{i,j}, \boldsymbol{c}) - \mu^{j}}{\sigma^{j}}$
                \STATE $A_k^{\text{mix}} \gets \sum_{j=1}^JA_k^{j}$ 
            \ENDFOR
            \STATE Compute GRPO loss $J(\theta)$
        \ENDFOR
        \STATE Update policy: gradient ascent on $J(\theta)$
   \ENDFOR
\end{algorithmic}
\end{algorithm}
\end{minipage}
}
\end{figure}

\section{Experiments}

\begin{figure*}[ht]
    \centering
    \includegraphics[width=\textwidth]{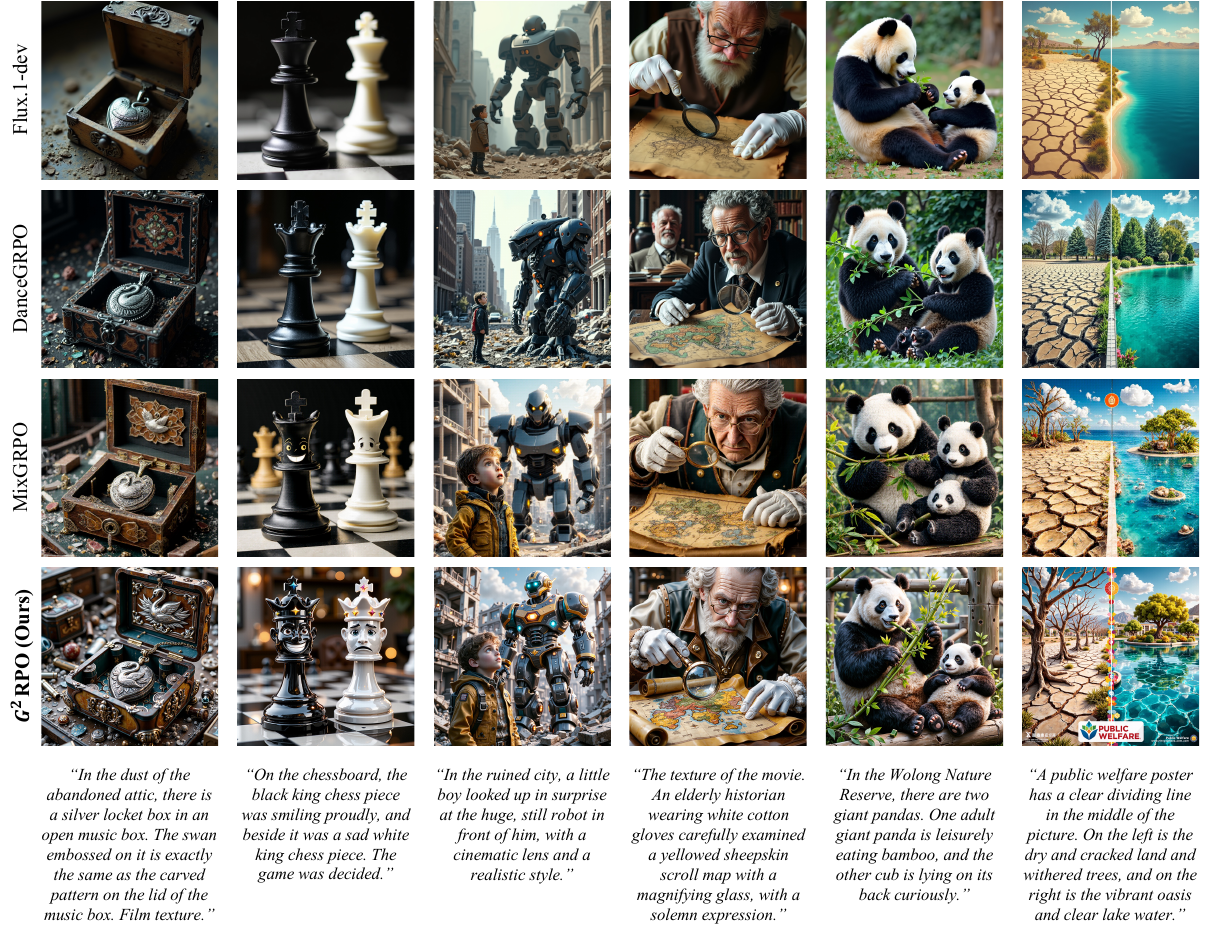}
    \caption{ \textbf{Qualitative Results.} Comparison with existing flow-based GRPO methods, in which our G$^2$RPO demonstrates superior performance in human preference alignment.
        }
    \label{fig:comparison}
\end{figure*}

\subsection{Implementation Details}

\textbf{Datasets and Backbone.} 
Following the same setting in 
DanceGRPO~\citep{xue2025dancegrpo} and MixGRPO~\citep{li2025mixgrpo}, 
the prompt dataset HPSv2~\citep{wu2023human} is utilized.
It contains 103,700 text prompts for training and 
400 diverse prompts for testing.
The text-to-image model employed for reinforcement learning is Flux.1-dev~\citep{flux2024},
a leading flow model in the community.

\noindent \textbf{Evaluation Metrics.} To evaluate the effectiveness and robustness of our G$^2$RPO,
multiple reward models are applied as evaluation metrics. 
These reward models assess the alignment between generated images and human preferences from multiple dimensions.
Specifically, HPS-v2.1 \textbf{(HPS)}~\citep{wu2023human}, CLIP Score \textbf{(CLIP)}~\citep{radford2021learning}, 
and Pick Score \textbf{(PS)}~\citep{kirstain2023pick} collectively assess semantic alignment and visual coherence. 
Image Reward \textbf{(IR)}~\citep{xu2023imagereward} focuses on visual quality and aesthetic appeal,
while Unified Reward \textbf{(UR)}~\citep{wang2025unified} is the SOTA
reward model that leverages the powerful visual understanding capabilities of LLM
comprehensively evaluates both alignment with the caption and overall quality.

\noindent \textbf{Evaluation Setting.}
Similar to DanceGRPO and MixGRPO, two experimental settings are employed. 
Firstly, a single HPS-v2.1 reward model is utilized for training
to verify the upper limit of improvement for \textit{in-domain performance},
i.e., both training and testing are conducted using the same reward model.
However, as demonstrated by DanceGRPO, 
HPS-v2.1 is prone to model hacking due to biases in the training set, 
leading to degradation in other evaluation metrics.
Similarly, our primary experiment involves joint training with both HPS-v2.1 and CLIP Score as reward models 
to acquire stable and robust results. 
Subsequently, we evaluate the model using \textit{out-of-domain metrics},
i.e., reward models that were not seen during training.

\noindent \textbf{Sampling Phase.}
Following DanceGRPO, a shared initialization noise is used
to generate a group of $12$ images from the same text prompt.
The total sampling step $T=16$ to enhance computational efficiency 
and the advantage clip $\varepsilon=5$ in Eq.~\ref{eq:f}.
The parameter $\eta=0.7$ in Eq.~\ref{eq:euler} directly determines the noise level,
which in turn defines the size of the stochastic exploration space in the SDE.
All comparison methods share the aforementioned parameter configuration.
For the Multi-Granularity Advantage Integration strategy, 
the set of distinct granularities $\Lambda=\{1,2,3\}$.

\noindent \textbf{Training Phase.}
All experiments are conducted using $16 \times$ NVIDIA H200 GPUs, with a batch size of $1$.
The AdamW optimizer is used, configured with a learning rate of $2 \times 10^{-6}$ and a weight decay of 
$1 \times 10^{-4}$. Mixed precision training is implemented using bfloat16 (bf16) format.

\begin{table}[!ht]
  \centering
  \caption{\textbf{Quantitative Results.} Comparison of results on various reward metrics.}
  \resizebox{\linewidth}{!}{

\begin{tabular}{llccccc}
\toprule
\textbf{Reward Model} & \textbf{Method} & \textbf{HPS} & \textbf{CLIP} & \textbf{PS} & \textbf{IR} & \textbf{UR} \\
\midrule
/ & Flux.1-dev & 0.305 & 0.388 & 0.226 & 1.040 & 3.621 \\
\midrule
\multirow{4}{*}{HPS}
& DanceGRPO & 0.353 & \textbf{0.375} & 0.228 & 1.233 & \textbf{3.548}\\
& MixGRPO & 0.378 & 0.358 & 0.225 & 1.266 & 3.421 \\
& \textbf{G$^2$RPO w/o MGAI} & 0.376 & 0.351 & 0.228 & 1.286 & 3.469 \\
& \textbf{G$^2$RPO} & \textbf{0.385} & 0.355 & \textbf{0.229} & \textbf{1.313} & 3.487 \\
\midrule
\multirow{4}{*}{\makecell{HPS \& CLIP}}
& DanceGRPO & 0.331 & 0.389 & 0.227 & 1.128 & 3.569 \\
& MixGRPO & 0.363 & 0.399 & 0.230 & 1.436 & 3.661 \\ 
& \textbf{G$^2$RPO w/o MGAI} & 0.372 & 0.395 & 0.234 & 1.421 & 3.688 \\
& \textbf{G$^2$RPO}  & \textbf{0.376} & \textbf{0.406} & \textbf{0.235} & \textbf{1.483} & \textbf{3.783} \\
\bottomrule
\end{tabular}
  }
  \label{tab:main_table}
\end{table}

\vspace{-2em}
\subsection{Main Results}
\textbf{Quantitative Evaluation.}
As shown in Tab.~\ref{tab:main_table}, DanceGRPO and MixGRPO 
serve as the baselines for comparison with our G$^2$RPO.
Flow-GRPO is not included since it employs a,  similar 
ODE-to-SDE conversion across all timesteps as contemporary DanceGRPO, but fit a specific reward target.
In addition, the Singular Stochastic Sampling strategy
without multi-granularity denoising (G$^2$RPO w/o MGAI) is also proposed.
It can be observed that when HPS-v2.1 is used solely as the training reward model,
our Singular Stochastic Sampling achieves a relative improvement of 6.52\% compared to DanceGRPO.
This indicates that constraining the stochasticity to a single step yields a precise reward signal, 
which in turn provides a more faithful optimization signal and
enables GRPO to enhance the optimization ceiling.
However, as demonstrated by DanceGRPO, 
optimizing solely with HPS-v2.1 can induce model hacking,
which in turn compromises other out-of-domain evaluation metrics.
Secondly, for the setting of multi-reward (HPS-v2.1 and CLIP Score) optimization,
the results indicate that G$^2$RPO outperforms other baselines
across both in-domain and out-of-domain rewards.
Under the multi-granularity denoising condition, 
groups of images at different granularities provide
a more comprehensive evaluation of the SDE sampling direction. 
This robust assessment facilitates improvements in out-of-domain metrics,
particularly on Unified Reward enhanced with LLM knowledge.

\begin{table}[!t]
  \centering
  \caption{\textbf{Ablation Study.} Comparison for different denoising granularities.}
  \resizebox{\linewidth}{!}{

\begin{tabular}{lcccccc}
\toprule
\textbf{Metrics} & \textbf{$\Lambda$} & \textbf{HPS} & \textbf{CLIP} & \textbf{PS} & \textbf{IR} & \textbf{UR} \\
\midrule
\multirow{4}{*}{\makecell{HPS \\ \& CLIP}}
 & $\{1\}$       & 0.372 & 0.395 & 0.234 & 1.421 & 3.688 \\
 & $\{1, 2\}$    & 0.375 & 0.404 & 0.234 & 1.468 & 3.759 \\
 & $\{1, 3\}$    & \textbf{0.378} & 0.404 & 0.234 & 1.465 & 3.760 \\
 & $\{1, 2, 3\}$ & 0.376 & \textbf{0.406} & \textbf{0.235} & \textbf{1.483} & \textbf{3.783} \\
\bottomrule
\end{tabular}
  }
  \label{tab:ablation_table}
\end{table}

\noindent \textbf{Qualitative Comparison.}
Fig.~\ref{fig:comparison} presents the qualitative comparison among the original Flux.1-dev,
DanceGRPO, MixGRPO, and our proposed G$^2$RPO. 
It can be observed that G$^2$RPO provides enhanced detail fidelity and
improves the consistency with the text prompt,
achieving superior alignment with human preferences.
For instance, in the second column, our G$^2$RPO faithfully captures the specified expressions 
and even the nuances of the chess pieces as described in the prompt, 
delivering finer details and higher visual quality.
Moreover, in the ``poster'' case depicted in the last column,
G$^2$RPO not only adheres to the spatial requirement of a clear left-right demarcation 
but also renders the reflections of the trees with remarkable clarity. 
The overall style of the image generated by G$^2$RPO is more consistent with the aesthetic demands of poster design.

\subsection{Ablation Study}
As described in Section~\ref {sec:method}, the Multi-Granularity Advantage Integration module integrates multi-granularity ODE sampling results
to evaluate the SDE sampling direction comprehensively. 
Different granularities represent diverse sampling intervals during denoising, 
controlled by the parameter set $\Lambda$.
To validate the effectiveness of multi-granularity fusion,
we perform ablation studies on various $\Lambda$ set shown in Tab.~\ref{tab:ablation_table}.
It can be found that as the number of selected granularities increases, the evaluation of each sampling direction becomes more comprehensive, 
facilitating a robust assessment through multi-granularity advantage fusion, 
which significantly improves performance across both
in-domain and out-of-domain reward models.

\begin{table}[!t]
  \centering
  \caption{Comprehensive evaluation of total denoising steps.}
  \resizebox{\linewidth}{!}{

\begin{tabular}{llccccc}
\toprule
\textbf{Inference Steps} & \textbf{Method} & \textbf{HPS} & \textbf{CLIP} & \textbf{PS} & \textbf{IR} & \textbf{UR} \\
\midrule
\multirow{4}{*}{10 Steps}
& Flux.1-dev & 0.289 & 0.388 & 0.225 & 0.939 & 3.504 \\
& DanceGRPO & 0.325 & 0.390 & 0.227 & 1.129 & 3.576\\
& MixGRPO & 0.358 & 0.401 & 0.230 & 1.431 & 3.641 \\
& \textbf{G$^2$RPO} & \textbf{0.378} & \textbf{0.408} & \textbf{0.235} & \textbf{1.519} & \textbf{3.805} \\
\midrule
\multirow{4}{*}{20 Steps} 
& Flux.1-dev & 0.300 & 0.389 & 0.226 & 1.034 & 3.575 \\
& DanceGRPO & 0.329 & 0.388 & 0.228 & 1.136 & 3.586\\
& MixGRPO & 0.363 & 0.401 & 0.230 & 1.430 & 3.651\\
& \textbf{G$^2$RPO}  & \textbf{0.376} & \textbf{0.407} & \textbf{0.235} & \textbf{1.511} & \textbf{3.806} \\
\bottomrule
\end{tabular}
  }
  \label{tab:more_exp}
\end{table}

\subsection{Further exploration of varying inference steps}
In the main experiment Tab.~\ref{tab:main_table}, 
we maintained the same inference settings as DanceGRPO and MixGRPO.
Notably, our MGAI module, which evaluates denoising samples of different granularities in a mixed manner, 
provides a more comprehensive assessment.
With the assistance of this module, G$^2$RPO exhibits stronger robustness to varying denoising step configurations.
As shown in Tab.~\ref{tab:more_exp}, all flow-based GRPO methods are jointly trained with HPS-v2.1 and CLIP. 
When the total inference timesteps are reduced to $20$ or even $10$ steps,
G$^2$RPO still achieved significant performance improvements across various in-domain and out-of-domain evaluation reward models,
thereby demonstrating the value of multi-granularity assessment.
\section{Conclusion}
This paper addresses the critical limitations of precisely evaluating the quality of 
denoising directions sampled by Flow-based GRPO for human preference alignment. 
We introduce Granular-GRPO (G$^2$RPO), a novel online RL framework that 
precisely localizes stochasticity to a single step within the denoising process 
and provides a comprehensive evaluation of SDE denoising directions 
by integrating the advantages derived from images at different denoising granularities. 
This innovative design enables the provision of dense, precise reward signals, 
thereby fundamentally improving optimization accuracy and leading
to a more robust and higher-quality alignment.
Our extensive experiments consistently demonstrate that G$^2$RPO achieves
superior performance across diverse reward conditions,
marking a significant advancement in aligning generative models 
with human preferences.

{
    \small
    \bibliographystyle{ieeenat_fullname}
    \bibliography{main}
}

\clearpage
\setcounter{page}{1}
\maketitlesupplementary
In the supplementary material,
we present additional implementation details (Section~\ref{sec:1}), 
additional quantitative comparison with baselines (Section~\ref{sec:2}),
visual ablation study of MGAI module (Section~\ref{sec:3}),
additional qualitative evaluation (Section~\ref{sec:4}), 
more visual samples of G$^2$RPO (Section~\ref{sec:5}),  
as well as the limitation of our method (Section~\ref{sec:6}), 
as a supplement to the main paper.

\section{Additional Implementation Details}\label{sec:1}
Tab.~\ref{tab:hyperparams} presents the detailed hyperparameter configuration used in our experiments. 
We keep the same hyperparameter configuration across all experiments.
\begin{table}[ht]
\caption{\textbf{Hyperparameter settings used in all experiments.}}
\vspace{-1em}
\centering
\resizebox{0.9\linewidth}{!}{%
\begin{tabular}{lclc}
\toprule
\textbf{Parameter} & \textbf{Value} & \textbf{Parameter} & \textbf{Value} \\
\midrule
Random seed & 42 & Learning rate & $2\times 10^{-6}$ \\
Train batch size & 1 & Weight decay & $1\times 10^{-4}$ \\
Warmup steps & 0 & Mixed precision & bfloat16 \\
Dataloader workers & 4 & Max grad norm & 1.0 \\
Eta & 0.7 & Sampler seed & 1223627 \\
Group size & 12 & Scheduler shift  & 3 \\
Sampling steps & 16  & Adv. clip max & 5.0 \\
Init same noise & Yes & Granularity $\Lambda$ & $\{1,2,3\}$ \\
The number of GPUs & 16 &  Clip range  & $1\times 10^{-4}$\\
\bottomrule
\label{tab:hyperparams}
\end{tabular}
}
\end{table}

\section{Additional Quantitative Evaluation}\label{sec:2}
To further demonstrate the robustness and superiority of G$^2$RPO over baseline methods,
as well as the comprehensive enhancement effects brought by multi-granularity evaluation to GRPO training, 
we employ the latest UniGenBench++~\cite{wang2025unigenbench++} as the benchmark for evaluation. 
UniGenBench++ is a unified and versatile benchmark for image generation 
that integrates diverse prompt themes with a comprehensive suite of fine-grained evaluation criteria. 
The benchmark encompasses 10 primary dimensions, 
covering semantic evaluation, image quality assessment, 
text alignment, and other aspects, 
to provide a complete and thorough evaluation of generative models.
Official offline evaluation model \textit{UniGenBench-EvalModel-qwen-72b-v1} is used as the VLM for evaluation.

As shown in Tab.~\ref{tab:unigenbench}, 
our Singular Stochastic Sampling strategy (i.e., G$^2$RPO without MGAI) 
enhances the precision of reward signals during GRPO training,
thereby aligning the model's output more closely with the reward models.
This leads to significant improvements in several evaluation metrics, including Attribute, Relation, and Text.
However, relying solely on single-granularity alignment 
tends to overfit to the biases inherent in the reward models. 
This causes the model to generate outputs that increasingly collapse into a narrower domain, 
resulting in degraded performance on metrics such as Style and Layout.

Notably, our Multi-Granularity Advantage Integration module (i.e., MGAI)
provides a more comprehensive evaluation of the sampling directions within a group,
enabling the selection of samples that exhibit advantages at both coarse-grained (structural) and fine-grained (textural) levels.
This leads to more robust model updates and effectively mitigates the risk of domain collapse.
Ultimately, our G$^2$RPO achieves substantial overall performance improvements on UniGenBench++.

\begin{table*}[!ht]
\centering
\caption{\textbf{Quantitative Comparison on UniGenBench~\citep{wang2025unigenbench++}}. Best scores are in \textbf{bold}, second-best in \underline{underlined}.} 
\resizebox{\linewidth}{!}{%
\begin{tabular}{lc|cccccccccc}
\toprule
\textbf{Model} & \textbf{Overall} & Style & World Know. & Attribute & Action & Relation. & Logic.Reason. & Grammar & Compound & Layout & Text \\
\midrule
Flux.1-dev        & 61.59   & \textbf{84.60} & 86.87 & 66.77 & 62.74 & 67.13 & 29.36 & \textbf{60.83} & 47.04 & 71.08 & 39.48 \\
DanceGRPO &66.06   & 76.00 & 87.82 & 73.72 & 68.82 & 75.38 & 38.76 & 59.63 & 64.95 & 80.78 & 34.77 \\
MixGRPO &\underline{66.60}  & \underline{80.80} & \underline{87.97} & 74.04 &\underline{68.82} & 75.00 & 38.99 & 59.49 & 64.82 & \textbf{81.34} & 34.77 \\
G$^2$RPO w/o MGAI  &66.32  & 73.70 & 86.08 & \underline{75.53} & 67.49 & \underline{77.28} & \underline{39.45} & \underline{60.70} & \underline{65.21} & 76.68 & \underline{41.09} \\
\textbf{G$^2$RPO} &\textbf{69.21}   & 76.20 & \textbf{89.08} & \textbf{79.91} & \textbf{71.10}& \textbf{78.17} & \textbf{42.66} & 58.82 & \textbf{70.36} & \underline{79.29} & \textbf{46.55} \\
\bottomrule
\label{tab:unigenbench}

\end{tabular} 
}
\end{table*}
\section{Visual Ablation Study of MGAI}\label{sec:3}
\begin{figure*}[ht]
    \centering
    \includegraphics[width=0.95\textwidth]{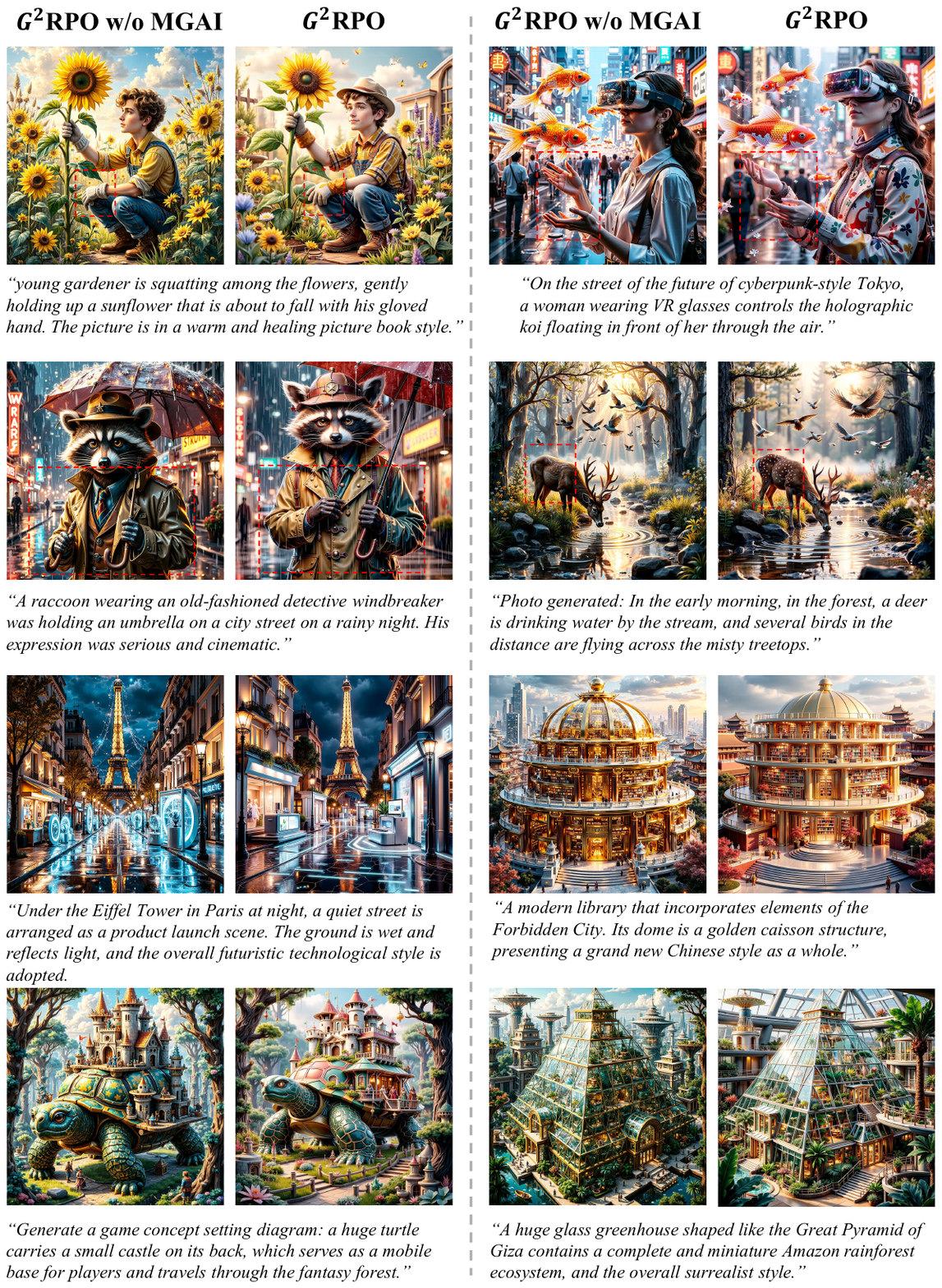}
    \vspace{-3em}
    \caption{ \textbf{Visual ablation study of MGAI module.}
        }
    \label{fig:ablation}
\end{figure*}

In this section, We conduct a visual ablation of the Multi-Granularity Advantage Integration (MGAI) module.
Fig.~\ref{fig:ablation} compares eight pairs of samples, with each pair sharing the same prompt and random seed.
Without MGAI, G$^2$RPO performs single-granularity denoising, 
whose trajectory is locked to a fixed step budget.
Group-wise advantage estimation under this setting is easily biased:
the reward model over-attends to fine details while ignoring coarse structural coherence,
so samples with higher reward frequently exhibit distorted textures or global misalignment.
Consequently, single-granularity generators tend to yield images that 
exhibit excessive textural detail while suffering from structural fragility.

The proposed MGAI alleviates this by re-weighting each group sample only 
when it simultaneously surpasses its peers at both coarse and fine scales,
forcing the policy to improve detail fidelity and global layout jointly.
VLM-centric metrics (Unified Reward~\cite{wang2025unified} and UniGenBench++)
capture this multi-dimensional gain, 
showing large boosts in prompt adherence, detail accuracy, and overall quality.
Conversely, uni-dimensional metrics such as HPS-V2.1~\cite{wu2023human},
CLIP Score~\cite{radford2021learning}, and Pick Score~\cite{kirstain2023pick} 
can already be over-fitted with our singular stochastic sampling strategy,
hence these metrics alone cannot fully capture the holistic gains conferred by MGAI.

\section{Additional Qualitative Comparisons}\label{sec:4}
\begin{figure*}[ht]
    \centering
    \includegraphics[width=0.95\textwidth]{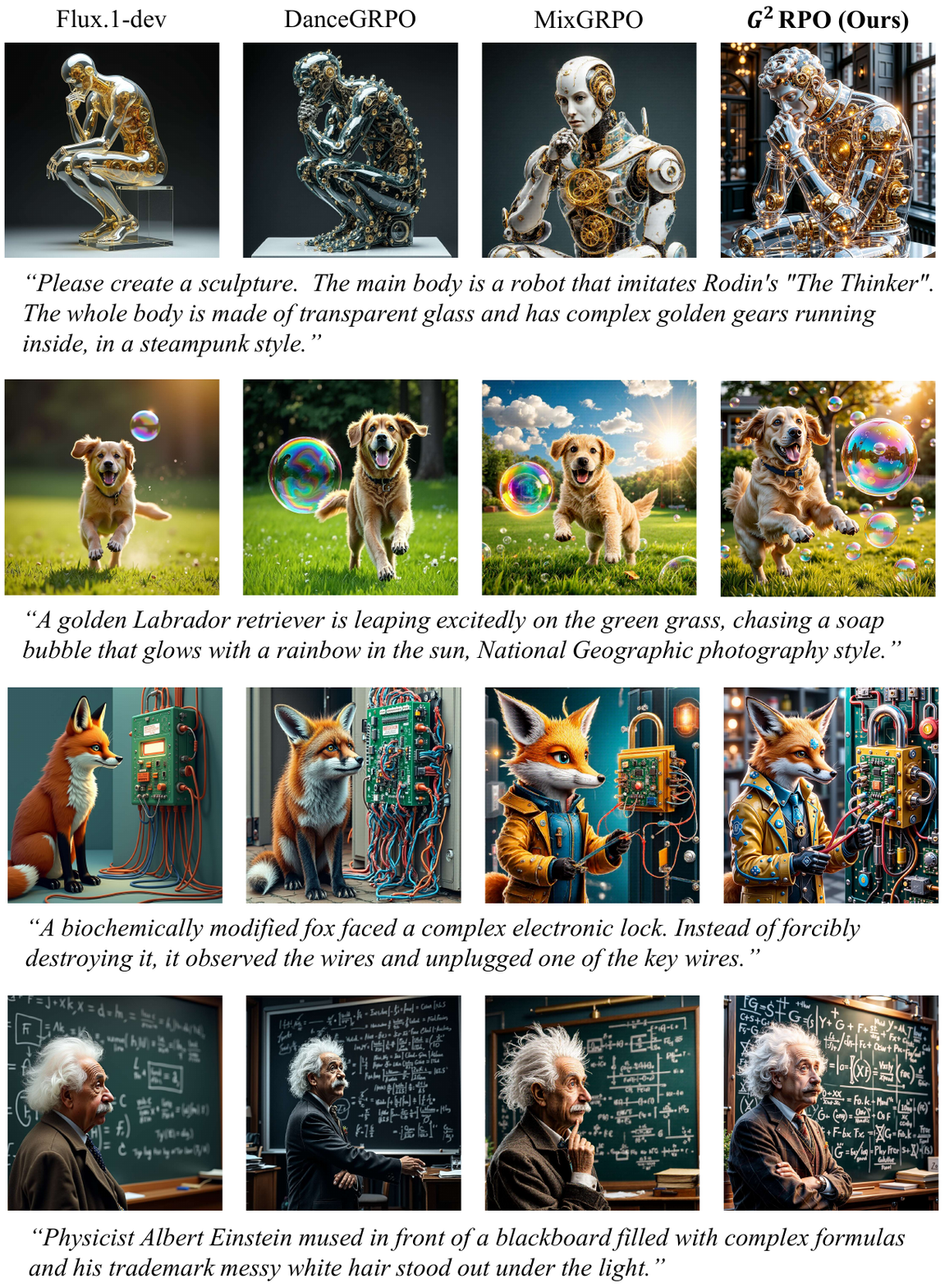}
    \vspace{-3em}
    \caption{ \textbf{Qualitative comparison with existing GRPO methods (1/3).}
        }
    \label{fig:comparison1}
\end{figure*}

\begin{figure*}[ht]
    \centering
    \includegraphics[width=0.95\textwidth]{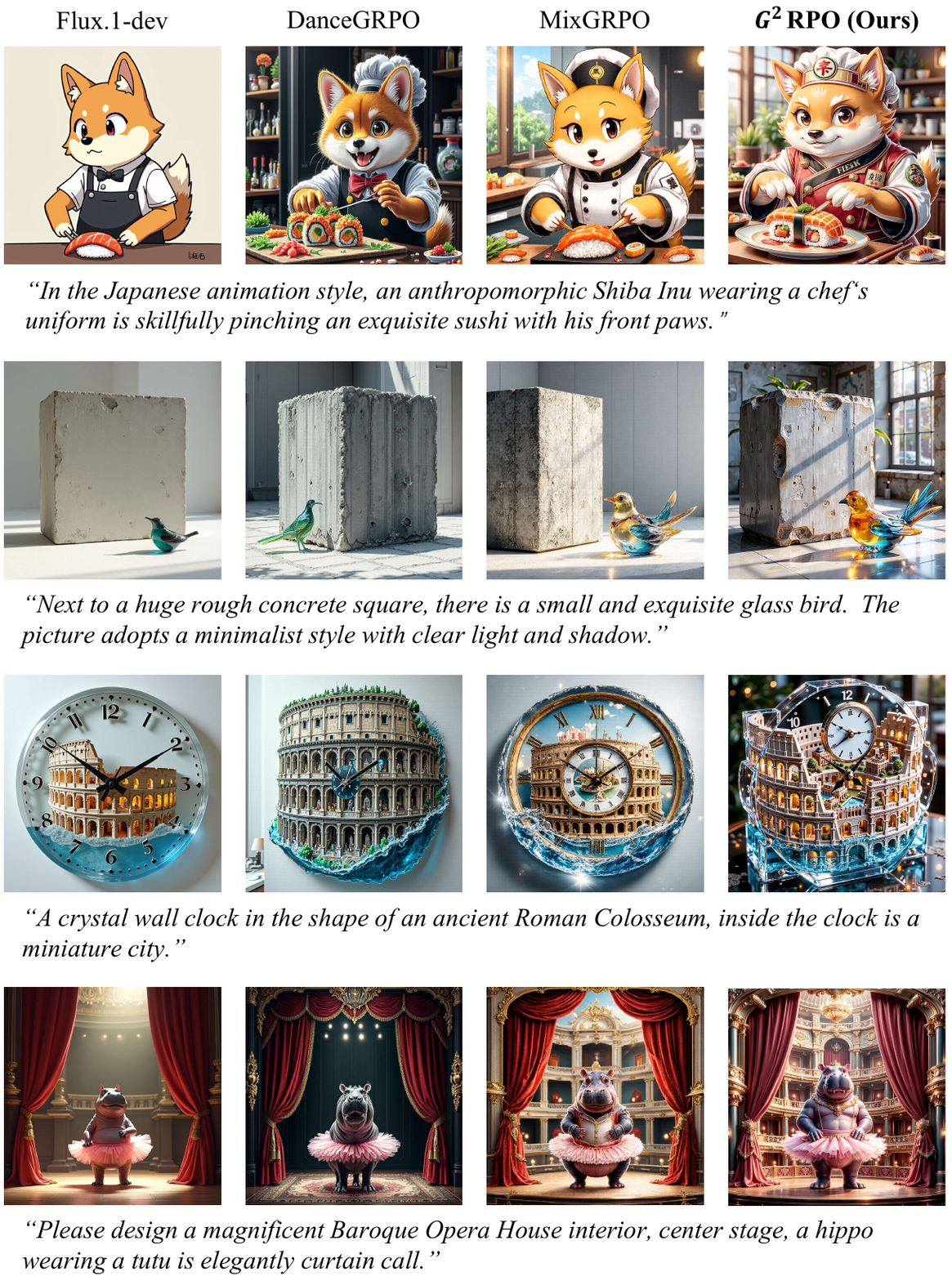}
    \vspace{-3em}
    \caption{ \textbf{Qualitative comparison with existing GRPO methods (2/3).}
        }
    \label{fig:comparison2}
\end{figure*}

\begin{figure*}[ht]
    \centering
    \includegraphics[width=0.95\textwidth]{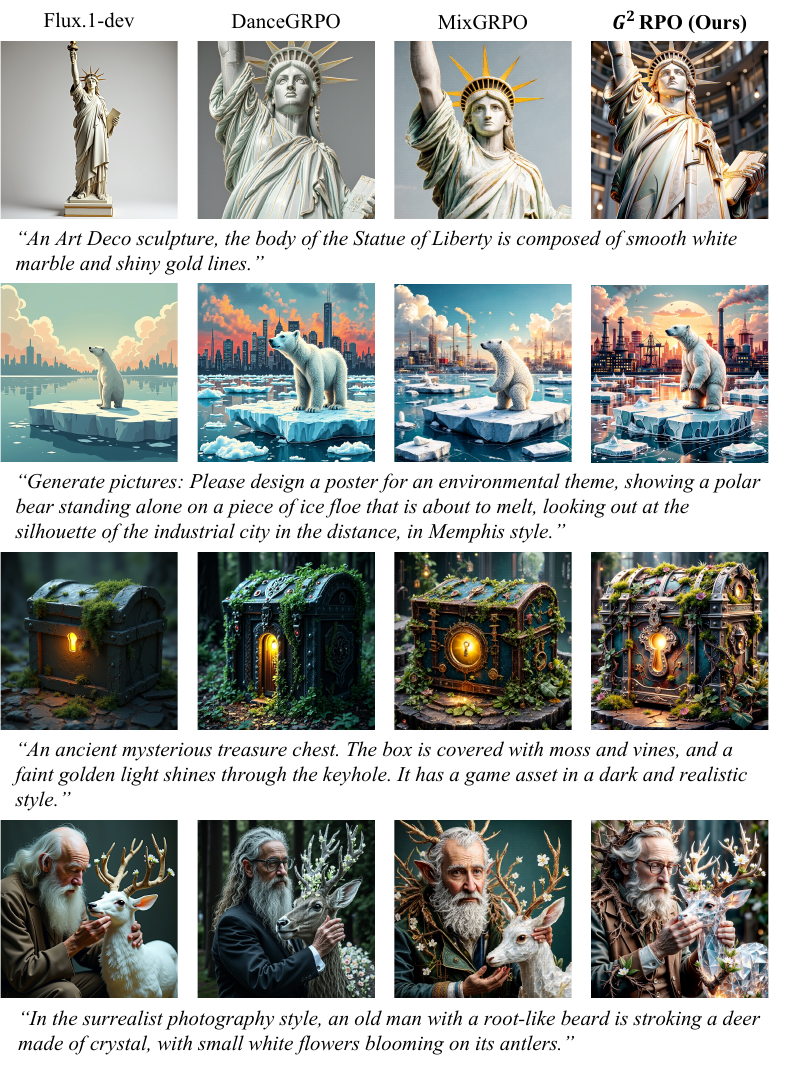}
    \vspace{-1.5em}
    \caption{ \textbf{Qualitative comparison with existing GRPO methods (3/3).}
        }
    \label{fig:comparison3}
\end{figure*}

In this section, we present more qualitative comparison results between our
G$^2$RPO and existing flow-based GRPO methods~\citep{xue2025dancegrpo,li2025mixgrpo}, as illustrated in Fig.~\ref{fig:comparison1}, 
Fig.~\ref{fig:comparison2}, and Fig.~\ref{fig:comparison3}.
G$^2$RPO achieves superior visual fidelity and text-image alignment.

\section{Gallery of G$^2$RPO}\label{sec:5}

\begin{figure*}[ht]
    \centering
    \includegraphics[width=0.95\textwidth]{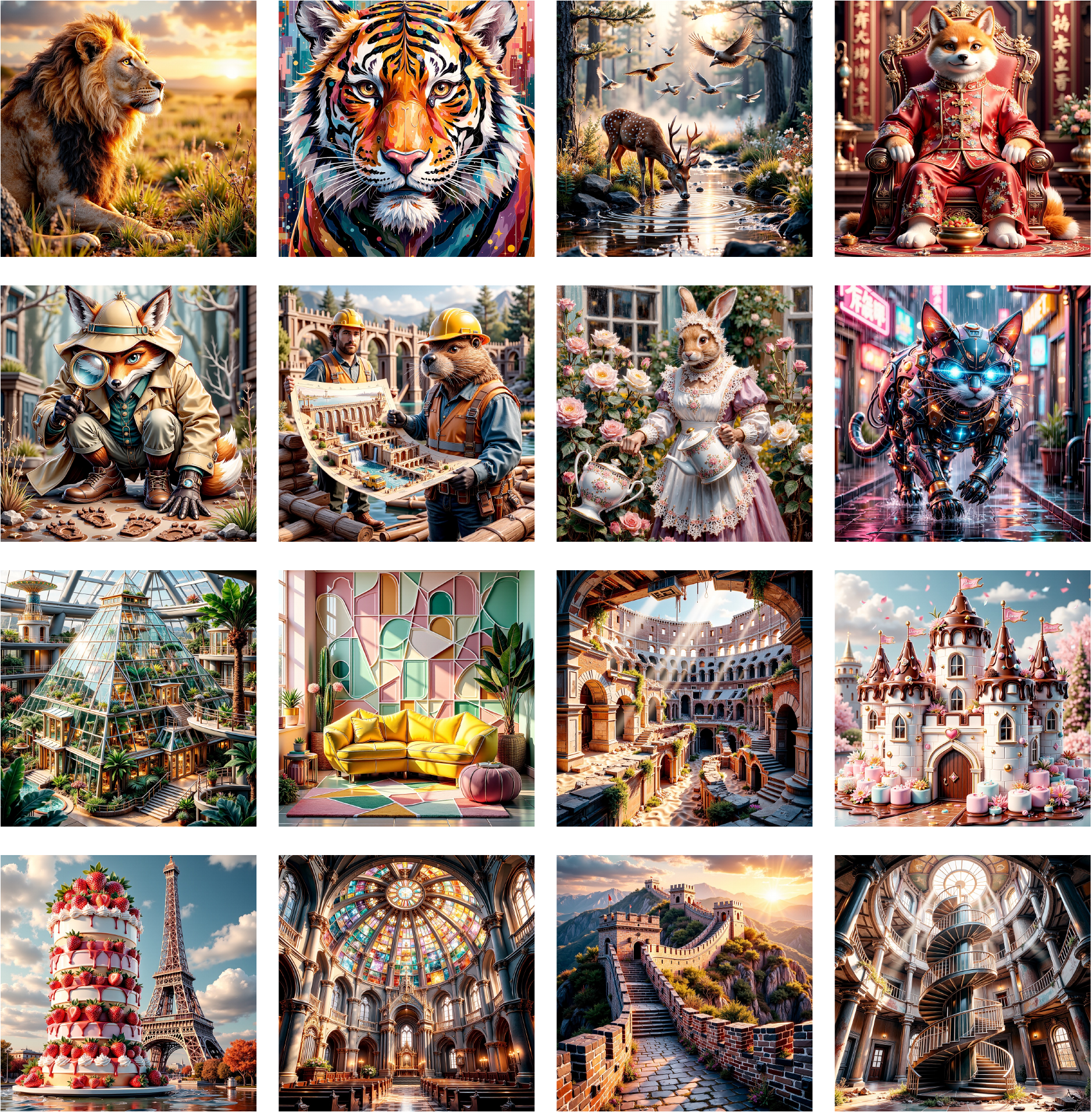}
    \caption{ \textbf{Gallery of G$^2$RPO (1/3).}
        }
    \label{fig:gallery1}
\end{figure*}

\begin{figure*}[ht]
    \centering
    \includegraphics[width=0.95\textwidth]{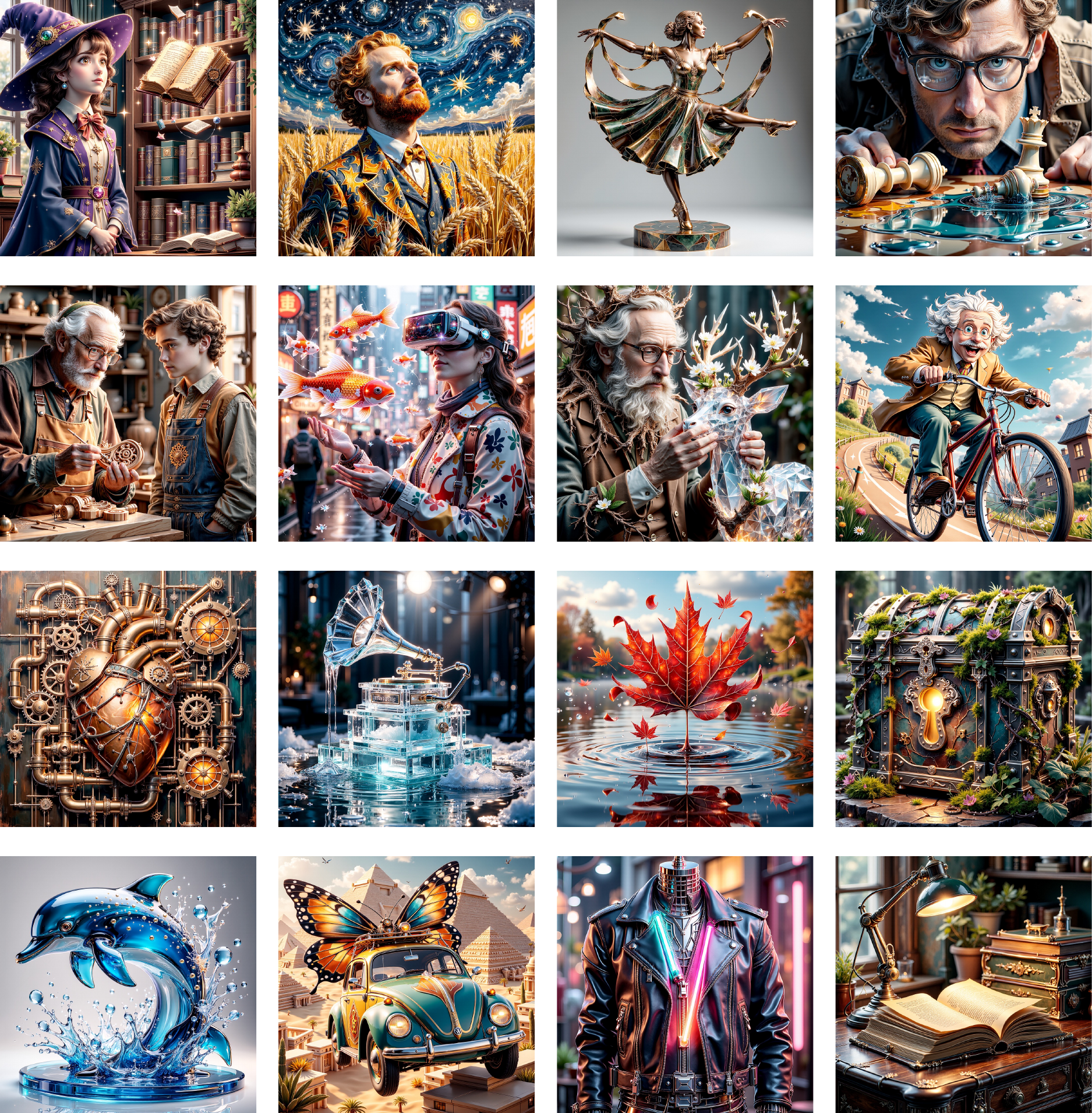}
    \caption{ \textbf{Gallery of G$^2$RPO (2/3).}
        }
    \label{fig:gallery2}
\end{figure*}

\begin{figure*}[ht]
    \centering
    \includegraphics[width=0.95\textwidth]{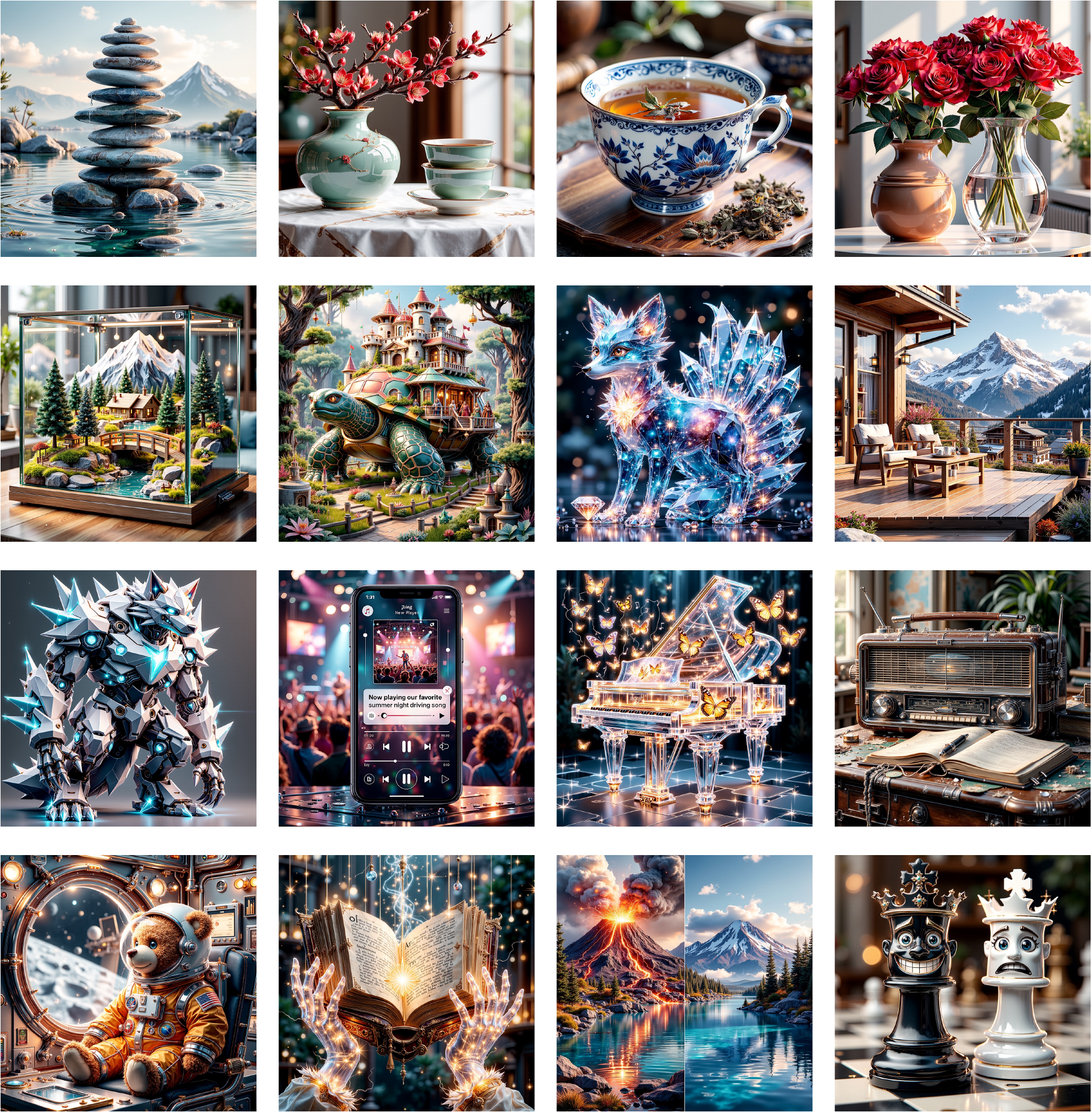}
    \caption{ \textbf{Gallery of G$^2$RPO (3/3).}
        }
    \label{fig:gallery3}
\end{figure*}

In this section, we provide more visual samples of the proposed G$^2$RPO to demonstrate its generation capability,
as shown in Fig.~\ref{fig:gallery1}, Fig.~\ref{fig:gallery2}, and Fig.~\ref{fig:gallery3}. 
Text prompts used to generate images are randomly sampled from UniGenBench++~\cite{wang2025unigenbench++}.

\section{Limitation}\label{sec:6}
Despite the advancements of our G$^2$RPO in human preference alignment,
it faces certain constraints. 
Specifically, G$^2$RPO incurs additional sampling time due to multi-granularity sampling.
We quantify the additional computational cost introduced by the granularity set $\Lambda=\{1,2,3\}$.
Let $\mathcal{M}=\left\{m_1, \ldots, m_k\right\}$ be the SDEs timesteps for GRPO training.
The standard single-granularity schedule requires 
$S_1=\sum_{m \in \mathcal{M}} m$ denoising steps.
Then, the tri-granular schedule introduces the additional step counts:
$$
S_2=\sum_{m \in \mathcal{M}}\left\lfloor\frac{m}{2}\right\rfloor, \quad S_3=\sum_{m \in \mathcal{M}}\left\lfloor\frac{m}{3}\right\rfloor .
$$
Total steps and relative overhead are therefore
$$
T_{\{1,2,3\}}=S_1+S_2+S_3, \quad \Delta_{\{1,2,3\}}=\frac{S_2+S_3}{S_1+S_2+S_3} .
$$
For example, with  $\mathcal{M}=\{16,15,...,9\}$, we have $T_{\{1,2,3\}} = 184$ and 
$\Delta_{\{1,2,3\}} \approx 45.7\%$.
Although our MGAI module adds a moderate training-time overhead,
this cost is incurred only once and leaves inference latency unchanged.
Meanwhile, as shown in Fig~\ref{fig:teaser} at the main paper, G$^2$RPO achieves
markedly faster alignment with the reward model compared with baseline method.

\newpage

\newpage

\newpage

\newpage

\newpage

\newpage

\newpage


\end{document}